\documentclass[letterpaper]{article} 
\usepackage{aaai2027}  
\usepackage[hyphens]{url}  
\usepackage{graphicx} 
\usepackage{natbib}  
\usepackage{caption} 
\usepackage{algorithm}
\usepackage{algorithmic}
\usepackage{algorithm}
\usepackage{algorithmic}
\usepackage{mdframed}
\usepackage{amsmath}
\usepackage{amssymb}
\usepackage{comment}

\DeclareMathAlphabet{\mathbbold}{U}{bbold}{m}{n}
\usepackage{newfloat}
\usepackage{listings}
\DeclareCaptionStyle{ruled}{labelfont=normalfont,labelsep=colon,strut=off} 
\floatstyle{ruled}
\newfloat{listing}{tb}{lst}{}
\floatname{listing}{Listing}

\usepackage{booktabs}
\nocopyright
\title{On Preference Coverage Collapse from Hindsight Relabeling in Multi-Objective Reinforcement Learning}
\title{On Preference Coverage Collapse from Hindsight Relabeling in Multi-Objective Reinforcement Learning}
\author {
    Baptiste Bonin\textsuperscript{\rm 1, \rm 2}\equalcontrib,
    Caro Strickland\textsuperscript{\rm 1, \rm 2}\equalcontrib,
    Audrey Durand\textsuperscript{\rm 1, \rm 2, \rm 3}
}
\affiliations {

    \makebox[0.2\textwidth][c]{\textsuperscript{\rm 1}Mila (Quebec AI Institute)}
    \makebox[0.2\textwidth][c]{\textsuperscript{\rm 2}Université Laval}
    \makebox[0.2\textwidth][c]{\textsuperscript{\rm 3}CIFAR AI Chair}

}

\begin{document}

\maketitle

\begin{abstract}
\begin{quote}
    
Hindsight relabeling which retroactively replacing a transition's goal with the outcome the agent actually achieved is an effective tool for improving sample-efficiency in Reinforcement Learning (RL). A natural extension to preference-conditioned multi-objective RL (MORL) relabels transitions with the preference direction the agent achieved rather than the one asked for. We show that this extension is frequently harmful: across four preference-conditioned off-policy algorithms spanning two critic backbones and two preference-sampling schemes on the continuous-control MO-Gymnasium suite, it degrades 19 of 36 algorithm-environment settings by as much as four standard deviations, improves only one, and leaves the rest unaffected.

The harm is not a symptom of noisy relabels; denoising the target recovers almost nothing, and neither prioritized sampling nor any buffer-structural choice reproduces it. Instead, repeated relabeling collapses the critic's coverage onto whatever narrow region of the preference space the agent happened to visit. We name this failure mode \emph{Preference Coverage Collapse}, and quantify it with abandoned preference mass (APM), a value-aware statistic that tracks the harm ($\rho = -0.73$) where a purely structural coverage count does not.

We then introduce \texttt{her\_mix}, a single-parameter convex combination pulling the achieved direction back towards the requested preference. At one fixed value across every algorithm and environment, it returns 16 of the 19 harmed settings to baseline, preserves and even improves the one setting in which relabeling helps, and cuts abandoned preference mass from $69\%$ to $6\%$. Protecting coverage over the preference simplex, not filtering noisy relabels, is what makes hindsight relabeling safe for MORL.
\end{quote}

\end{abstract}


\section{Introduction}

Although Reinforcement Learning (RL) has been shown to be effective across a range of reward-driven problems \cite{mnih2015human,duan2016benchmarking,haarnoja2018soft}, many RL algorithms require large amounts of training data and remain especially sample-inefficient in the presence of sparse or delayed rewards. In Multi-Objective Reinforcement Learning (MORL), this inefficiency is compounded. Rather than learning a single behavior, a preference-conditioned agent must learn to cover an entire space of trade-offs from the same finite experience pool, effectively multiplying the amount it needs to learn without a corresponding increase in data \cite{yang2019generalized,abels2019dynamic}. 
Hindsight Experience Replay (HER) \cite{andrychowicz2017hindsight}, introduced for sparse-reward goal-conditioned RL, is among the most effective remedies.
Rather than discarding a trajectory that fails to reach its assigned goal, HER relabels it with the outcome the agent actually achieved and retrains on it, turning failures into usable learning signal. The substitution costs nothing because
a goal and an achieved outcome are the same kind of object
: whatever the agent reached is itself a perfectly valid goal in hindsight.


The same principle admits an analogue in preference-conditioned MORL, 
where the policy is conditioned on a preference vector rather than a goal. Instead of substituting an achieved state for the goal, the agent substitutes the trade-off it actually achieved for the preference it was given.
This substitution has already been adopted in prior work \cite{basaklar2022pd,shianifar2026hindsight} (discussed in Section~\ref{sec:related_work_HER_in_MORL}), justified by its relationship to HER rather than by independent analysis.



\begin{figure}[htpb!]
\centering
\includegraphics[width=\linewidth]{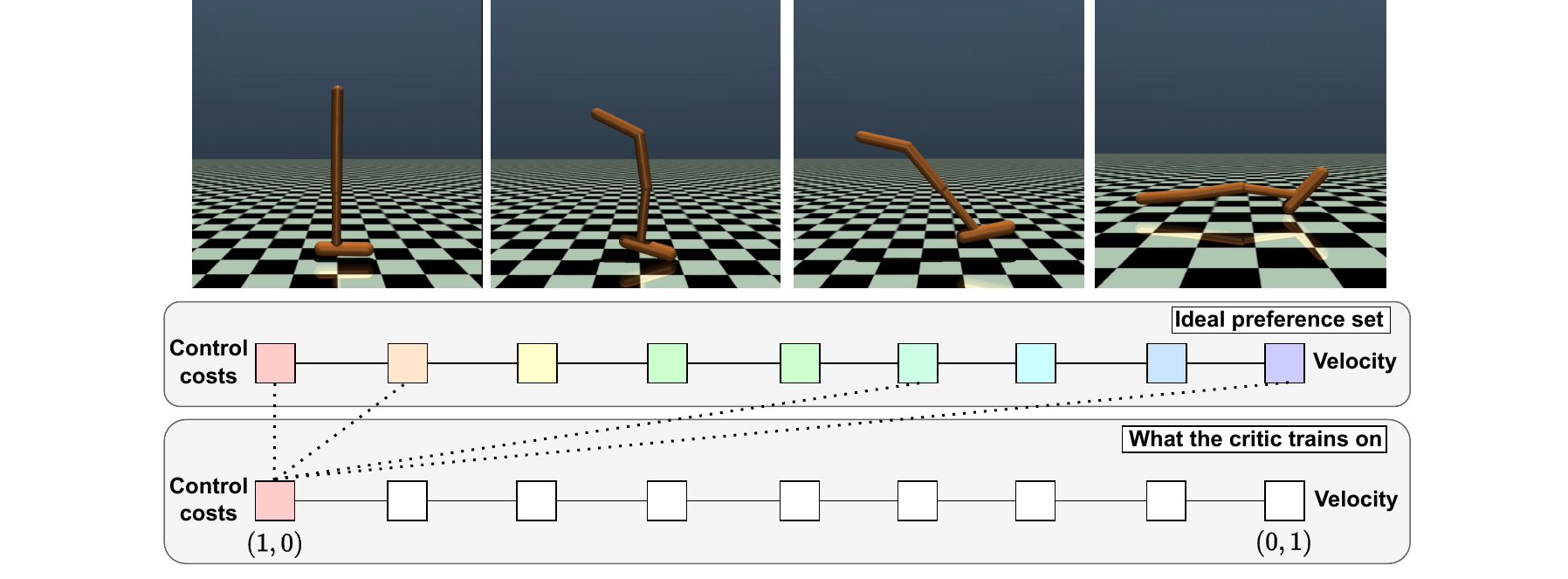}
\caption{\textbf{A single training episode collapses the preferences a policy trains on.} A
\texttt{hopper} agent falls within a few dozen steps; relabeling substitutes that narrow achieved
trade-off for every preference requested during the episode. Repeated, the critic sees a cluster of
achieved directions instead of the full requested range.}
\label{fig:hopper_example}
\end{figure}

The case for relabeling in preference-conditioned MORL is, at first glance, strong:
relabeling promises to multiply off-policy data for free, requiring no additional environment interaction and no architectural changes. Yet, the precondition behind HER's substitution does not obviously carry over. 
Rewards in MORL 
are typically dense and continuous-valued, and not every objective contains a natural notion of \emph{achievement} in the way a goal is achieved (a cost objective, for instance, is never reached so much as minimized). Consider a hopper agent asked to trade off forward speed against energy (control costs) that falls over after some small number of timesteps. The trade-off it ``achieved'' is whatever brief forward shuffle it managed before collapsing, which says little about the speed versus energy tradeoff it was asked for. Relabeling nonetheless trains the critic to treat that shuffle as the preference the agent wanted, and repeating this across many short episodes teaches the critic to value only the narrow corner it happened to reach 
(Figure~\ref{fig:hopper_example}). 
The technique adopted to fill an agent's coverage of the preference space instead collapses it. A failure mode we name \emph{Preference Coverage Collapse} (PCC).

Whether the substitution remains well-defined once these structural differences are accounted for has not been directly tested. As we show, the answer determines not only whether relabeling helps or hinders learning in MORL, but how it can be made safe when it doesn't. We therefore ask:

\begin{mdframed}
\center``does HER-style preference relabeling improve off-policy MORL, where does it fail, why, and can it be repaired?''
\end{mdframed}

The answer is sharper than current practice assumes: relabeling is harmful in the majority of settings where it has any effect at all. Our primary contributions are as follows:

\begin{enumerate}
    \item \textbf{An empirical study of preference relabeling in off-policy MORL.} We provide the first controlled evaluation of whether HER-style preference relabeling helps or harms off-policy MORL performance across four preference-conditioned algorithms spanning two critic backbones and two preference-sampling schemes. It degrades 19 of 36 algorithm-environment settings by as much as four standard deviations and improves exactly one.

    \item \textbf{A hidden failure mode in relabeling design, and its resolution.} We identify a clipping artifact in the standard relabeling scheme that causes it to silently ignore cost objectives, and show that a sign-robust fix resolves most of the apparent harm previously attributed to relabeling on cost-objective environments.

    \item \textbf{A mechanism for the harm, and a value-sensitive diagnostic for it.} We show that the remaining harm is not explained by noisy relabeling targets, prioritization, or buffer structure, but rather by a collapse in the critic's coverage over the preference simplex. We introduce Abandoned Preference Mass (APM),  which tracks the harm far more tightly ($\rho = -0.73$) than a purely structural count of the trade-offs a policy offers ($+0.40$), because it is value-aware.

    \item \textbf{A fix that generalizes without tuning.} We introduce \texttt{her\_mix}, a single-parameter interpolation between the achieved and requested preference. Rolled out untuned at one fixed coefficient across all 36 settings. It recovers 16 of the 19  harmed settings, preserves and amplifies the one setting where relabeling helps and leaves the neutral settings unaffected.

    \item \textbf{A negative result on replay-buffer structure.} We show that storage, partitioning, eviction, and capacity choices produce no harm at 16-seed power, isolating relabeling as the buffer design axis that degrades performance.
\end{enumerate}

\section{Background}
\label{sec:background}

\subsection{Reinforcement Learning}

We use the standard RL formulation \cite{sutton1998reinforcement}: a fully observable environment $(\mathcal{S}, \mathcal{A}, r, p, \gamma)$ with $r : \mathcal{S} \times \mathcal{A} \rightarrow
\mathbb{R}$ and $\gamma \in [0,1)$, a policy $\pi(a|s)$, discounted return $G_t$, and action-value function $Q^\pi(s,a) = \mathbb{E}_\pi[G_t \mid s_t=s, a_t=a]$.


\subsection{Multi-Objective Reinforcement Learning}

In MORL, the scalar reward is replaced by a vector-valued reward function $\mathbf{r} : \mathcal{S} \times \mathcal{A} \rightarrow \mathbb{R}^d$, where $d$ is the number of objectives. MORL algorithms can be broadly divided into multi-policy approaches, which learn a set of policies to jointly approximate the Pareto front, and single-policy (preference-conditioned) approaches, which learn a single policy that adapts its behavior at inference time based on a preference vector \cite{roijers2013survey}. We focus on the latter.

The agent is conditioned on a preference vector $\mathbf{w} \in \mathcal{W}$, typically constrained to the $(d-1)$-simplex specifying a relative weighting over objectives. Behavior is evaluated by a scalarized return, commonly the linear scalarization $\mathbf{w}^\top G_t$ of the vector-valued discounted return $G_t \in \mathbb{R}^d$, yielding the preference-conditioned action-value function $Q^\pi(s, a, \mathbf{w}) = \mathbb{E}_\pi\big[\mathbf{w}^\top G_t \mid s_t = s, a_t = a\big].$

\subsection{Evaluation Metrics}

Because MORL algorithms are generally evaluated on how well they recover the Pareto front rather than a single scalar return \cite{hayes2022practical}, we adopt hypervolume and expected utility as our primary evaluation metrics.

\textbf{Hypervolume (HV).} measures the volume of objective space dominated by a set of achieved returns relative to a reference point $\mathbf{r}_{\text{ref}}$, rewarding both the extent and spread of the recovered front; it is, however, sensitive to choice of $\mathbf{r}_{\text{ref}}$.

\textbf{Expected Utility Metric (EUM).} Is the expected scalarized return under a preference distribution $p(\mathbf{w})$ \cite{zintgraf2015quality,hayes2022practical}.
$\text{EUM} = \mathbb{E}_{\mathbf{w} \sim p(\mathbf{w})}\left[\, \mathbf{w}^\top G^{\pi_\mathbf{w}} \,\right], $
where $G^{\pi_\mathbf{w}}$ is the return achieved by the policy conditioned on $\mathbf{w}$. Unlike HV, EUM requires no reference point $\mathbf{r}_{\text{ref}}$ and scales directly with the objectives the agent is optimizing for under each sampled preference.


\textbf{Effect size.} We report standardized effect sizes as Cohen's $d$ \cite{cohen2013statistical}, the difference in means between an intervention and baseline divided by their pooled standard deviation, $d = \frac{\bar{x}_{\text{treat}} - \bar{x}_{\text{base}}}{s_{\text{pooled}}}.$

\subsection{Replay Buffers and Prioritized Experience Replay}

Off-policy actor-critic algorithms store past transitions in a \emph{replay buffer} $\mathcal{D}$, a set of tuples $(s, a, \mathbf{r}, s', \mathbf{w})$ collected from interaction with the environment, and train by repeatedly sampling minibatches from $\mathcal{D}$ rather than using only recent experience. The standard buffer samples transitions uniformly at random, implicitly treating each stored transition as equally useful for learning \cite{lillicrap2019continuouscontroldeepreinforcement}.

Prioritized Experience Replay (PER) \cite{schaul2015prioritized} relaxes this by sampling transitions with probability proportional to $|\delta_i|^{\alpha}$, where $\delta_i$ is the TD error and $\alpha$ controls the strength of prioritization. It is important to note that PER changes only how often a transition is sampled, leaving the transition's content completely unmodified.

\subsection{Hindsight Experience Replay}
\label{sec:her}

Whereas PER changes how \emph{often} a transition is sampled, HER changes what a transition \emph{says}. HER was introduced for goal-conditioned RL under sparse, binary rewards,  $r(s, a, g) = \mathbbold{1}[\|s' - g\| < \epsilon]$: an agent rarely reaches its assigned goal early in training, so most episodes yield an uninformative zero reward regardless of how transitions are sampled or prioritized. HER addresses this by relabeling a failed transition's goal post hoc. Given the transition $(s, a, s', g)$, it additionally stores $(s, a, s', g')$ for some state $g'$ achieved later in the same trajectory, turning a failure in a successful and informative transition.

This substitution is valid because a goal and an achieved outcome are both elements of $\mathcal{S}$, and success is defined purely by proximity to whichever one is supplied. Thus, swapping $g$ for $g'$ leaves the reward function's definition intact. This property, rather than the relabeling procedure itself, is what our results explore once the same substitution is applied to preference vectors $\mathbf{w}$ in place of goals $g$.

\subsection{Preference Relabeling in MORL}

Extending HER's substitution to the preference-conditioned setting means replacing the relabeled goal $g'$ with a relabeled preference $\mathbf{w}'$. For a stored transition $(s, a, \mathbf{r}, s', \mathbf{w})$, the natural parallel to the ``the goal actually reached'' is the objective vector the agent actually achieved along the trajectory (the per-step reward vector $\mathbf{r}$, or an aggregate such as return-to-go). We refer to this achieved objective vector as $\hat{\mathbf{o}}$. The default operator, \texttt{her\_achieved}, maps $\hat{\mathbf{o}}$ onto the $(d-1)$-simplex $\mathcal{W}$ via $\mathbf{w}' = \text{normalize}(\text{clip}(\hat{\mathbf{o}}, 0, \infty))$,  $\text{normalize}(\mathbf{x}) = \mathbf{x}/\sum_i x_i$, clipping negative entries to zero so $\mathbf{w}'$ is valid; the relabeled transition $(s, a, \mathbf{r}, s', \mathbf{w}')$ is stored alongside the original.

This differs from HER's original setting in two ways. First, HER solves a sparse-reward problem, whereas MORL's scalarized reward $\mathbf{w}^\top \mathbf{r}$ is generically nonzero for any $\mathbf{w}$: the motivation here is data reuse, not signal recovery. Second, and more consequentially, HER's validity argument (Section~\ref{sec:her}) does not transfer. It relied on
success being defined by proximity to whichever goal is supplied, which makes any achieved state an automatically valid stand-in. A preference is never ``reached'' in that sense: $\hat{\mathbf{o}}$ is
a noisy trade-off realized on one trajectory, not an outcome the agent aimed for, and nothing guarantees that $\mathbf{w}'$ corresponds to a trajectory the agent would produce if it optimized for $\mathbf{w}'$ from the start.

The clip is also silently ill-posed for objectives that are negative by construction: if
$\hat{o}_i < 0$ on effectively every transition then $\text{clip}(\hat{o}_i, 0, \infty) = 0$ always,
and $\mathbf{w}'$ can never place weight on objective $i$ whatever the trajectory's true trade-off. Section~\ref{sec:results} quantifies this and evaluates a sign-robust alternative, \texttt{her\_scaled}, based on per-objective min-max normalization.

\section{Related Work}

\subsection{HER in single-objective RL}

As discussed in Section \ref{sec:background}, HER relabels failed goal-conditioned trajectories with achieved outcomes; the original paper found that relabeling toward states from the trajectory's near future consistently outperformed alternative relabeling targets, which the authors attributed to the relevance of near-future states as possible goals \cite{andrychowicz2017hindsight}. Subsequent work has continued to treat the relabeling target as a nontrivial design choice. HER is known to perform poorly when the assigned goal is far from the agent's initial state \cite{ren2019exploration}, motivating curriculum-based extensions that prioritize intermediate goals \cite{fang2019curriculum}. Even where the achieved-outcome substitution is valid by construction, \emph{which} outcome to relabel toward, and under what conditions, impacts its benefit. As we discuss, this scrutiny has not carried over to MORL's adoption of the same substitution.

\subsection{HER-style relabeling applied to MORL}
\label{sec:related_work_HER_in_MORL}

The single-objective substitution principle has been carried over to preference-conditioned MORL. PD-MORL \cite{basaklar2022pd} adopts a HER-inspired buffer for both its MO-DDQN-HER and MO-TD3-HER variants, explicitly following \cite{andrychowicz2017hindsight}; however, its relabeling draws preferences independent of the trajectory, rather than substituting the achieved trade-off itself, and as such does not test the substitution HER's original motivation actually implies. More recently, Hindsight Preference Replay (HPR) \cite{shianifar2026hindsight} implements the achieved trade-off substitution directly for Concave-Augmented Pareto Q-Learning (CAPQL) \cite{lu2023multi}, discussed further in Section \ref{subsec:algorithms_and_environments}, motivating the approach as a direct generalization of the HER principle into preference space. In both cases, adoption is motivated by HER's single-objective success rather than by analysis of whether HER's underlying precondition, that an achieved outcome and a requested goal are directly substitutable, holds in the preference-conditioned multi-objective setting.

\subsection{Off-policy MORL algorithms}
\label{sec:off_policy_MORL_algs}

Off-policy MORL algorithms fall broadly along two main design axes: the underlying critic/update rule inherited from single-objective RL, and the scheme used to sample or condition on preferences during training \cite{lu2023multi}. Early off-policy approaches extended DQN-style value learning to multiple objectives through scalarized targets \cite{mossalam2016multi}, building on scalarization techniques developed for tabular multi-objective Q-learning \cite{van2014multi}. More recent work has moved toward continuous control backbones (SAC, TD3) paired with preference-conditioned critics capable of representing an entire Pareto front within a single policy \cite{yang2019generalized}. Preference-sampling schemes vary similarly, with some methods drawing preferences uniformly from the $d$-simplex during training \cite{basaklar2022pd} while others bias sampling toward regions expected to be more informative, such as cone-based sampling around directions that the policy is already competent in \cite{lu2023multi}, or Generalized Policy Improvement (GPI) over a maintained set of preference-conditioned policies \cite{alegre2023sample}.

\subsection{Diversity metrics for coverage}

Measuring how evenly a set of samples is distributed, rather than only their quantity, is a long-studied problem, where Hill numbers \cite{hill1973diversity} and related entropy-based diversity indices \cite{jost2006entropy} quantify the ``effective number'' of categories represented in a population, collapsing to the raw category count under a uniform distribution and shrinking as mass concentrates. We adapt the same construction to the preference simplex, treating each relabeled preference as a sample and asking how many effectively distinct regions of the simplex it covers. 
Diversity-style objectives also appear within RL, in quality-diversity, novelty-search and
skill-discovery methods \cite{lehman2011abandoning,pugh2016quality,eysenbach2018diversity}.

Coverage collapse in preference-conditioned MORL specifically has also been addressed directly at the architecture-level, independent of the replay buffer. \citeauthor{kubo2026single} show that under Smooth Tchebycheff (STCH) scalarization, the map from preference to optimal objective vector is Lipschitz continuous, and they derive an update rule (CMDPI) that exploits this for better preference-space coverage. Conversely, D$^3$PO \cite{ambadkar2026preference} introduces a diversity-driven regularizer that directly penalizes representational collapse across the preference simplex. These approaches prevent collapse structurally, at the level of the scalarization or loss function. Whether a collapse-resistant architecture such as D$^3$PO remains robust once \texttt{her\_achieved}-style relabeling is layered on top is, to our knowledge, untested.

Our use of a diversity metric differs in purpose; rather than optimizing for diversity, or building an architecture robust to its loss, we use it purely diagnostically, to characterize a change in what a fixed training procedure covers as a side effect of relabeling.

\section{Preference Coverage Collapse}
\label{sec:pcc}

\subsection{Coverage}
\label{subsec:metrics}

\emph{Preference Coverage Collapse} (PCC) is the narrowing of the trade-offs a policy actually
serves under achieved-preference relabeling: a covering policy answers different regions of
$\mathcal{W}$ with different points of its achieved return set $\mathcal{F}$, and a collapsed policy
serves the whole simplex from one corner. We quantify this in two ways.


\noindent \textbf{Effective coverage (C).} For each $\mathbf{w} \in W$, we let $p_k \in \mathcal{F}$ be the maximizing return and $\rho_k$ the share of $W$ for which $p_k$ is optimal. $C$ here is the exponential Shannon entropy of this distribution (a Hill number giving the effective number of distinct trade-offs offered):


\begin{equation}
\label{eq:coverage-number}
    C(\mathcal{F}) = \exp\left(-\sum_{k=1}^{|\mathcal{F}|} \rho_k \ln \rho_k\right).
\end{equation}

$C \in [1, |\mathcal{F}|]$: $C \approx 1$ under total collapse (one return optimal almost everywhere), $C$ approaches $|\mathcal{F}|$ when every front point serves a distinct region of $W$. $C$ is scale-invariant and comparable across environments.



\noindent \textbf{Abandoned Preference Mass (APM).} $C$ is purely structural and cannot distinguish healthy specialization from uniform failure. APM complements $C$, measuring the fraction of $W$ where utility falls meaningfully below a baseline front: 


\begin{equation}
\mathrm{APM}(\mathcal{F};\mathcal{F}_0) \;=\; \frac{|\mathcal{A}_\tau|}{|W|},
\label{eq:apm}
\end{equation}
where $\mathcal{A}_\tau=\{\mathbf{w}\in W: U_{\mathcal{F}}(\mathbf{w})<(1-\tau)\,U_{\mathcal{F}_0}(\mathbf{w})\}$
is the set of abandoned preferences and $\tau$ sets what counts as meaningful; we use $\tau = 0.1$ throughout, so a preference is abandoned when the policy realises less than $90\%$ of the baseline's utility there. APM lies in $[0,1]$ and equals $0$ for the baseline itself; unlike $C$ it is value-aware, reading near $0$ wherever the policy matches or beats the baseline.

Both are functionals of the same realised fronts as EUM, so APM recasts the harm as lost coverage rather than independently explaining it; the causal account lives at the training level and is established empirically in Section~\ref{subsec:mechanism}.


\subsection{Mixed Relabeling}
\label{subsec:hermix}

If the failure mode is a collapse of the conditioning distribution, the repair is to keep that distribution spread out.\texttt{her\_mix} blends, per transition, the achieved direction with the originally collected preference:

\begin{equation}    
\mathbf{w}_{\text{used}} = (1 - \lambda)\, \mathbf{w}_{\text{collected}} + \lambda\, \mathbf{w}_{\text{achieved}},
\label{eq:hermix}
\end{equation}

a convex combination of two points of $\mathcal{W}$ and therefore itself a valid preference. It adds one hyperparameter and one line of code; we fix $\lambda = 0.25$ everywhere, with no per-setting tuning. As a falsification control, we also evaluate a TD-gated variant that rejects relabels above a running TD-error percentile but substitutes fully on the transition it accepts. If the harm is coverage loss rather than noisy outliers, gating should underperform mixing.


\section{Experiments}

\subsection{Algorithms and environments}
\label{subsec:algorithms_and_environments}

We evaluate four preference-conditioned off-policy algorithms, chosen to span the two design axes
of Section~\ref{sec:off_policy_MORL_algs} rather than to represent them exhaustively.
\textbf{CAPQL} \cite{lu2023multi} pairs a SAC \cite{haarnoja2018soft} backbone with an angle-cone
preference sampler; \textbf{GPI-PD} \cite{alegre2023sample} pairs TD3 \cite{fujimoto2018addressing} with GPI weight-support sampling. They differ on both axes at once,
so an effect that reproduces across both is unlikely to be an artifact of either choice. The two ablations swap one component each: \textbf{MO-TD3} is TD3 with the cone sampler, \textbf{CAPQL-uniform} is SAC with uniform simplex sampling: completing a $2\times2$ grid over
$\{\text{update rule}\} \times \{\text{sampler}\}$ that separates an effect tracking the backbone
from one tracking the sampler. Appendix A of the supplementary material gives the full descriptions. All four run with library-default hyperparameters; the replay buffer is the only component that varies.

The benchmark is the continuous-action MO-Gymnasium suite \cite{felten_toolkit_2023}: eight MO-MuJoCo v5 tasks (\texttt{hopper-2obj}, \texttt{hopper}, \texttt{walker2d}, \texttt{halfcheetah}, \texttt{swimmer}, \texttt{ant-2obj}, \texttt{ant}, \texttt{humanoid}) and \texttt{mountaincar-continuous}, with two or three objectives each\footnote{\texttt{mo-reacher-v5} is excluded due to its discrete action space.}. Crossing algorithms with environments yields 36 settings, the unit of analysis below.

\subsection{Protocol}
Every run trains for 150k environment steps with evaluation every 15k. Final EUM is the value at the
last evaluation point. Comparisons that support a claim use 16 random seeds per setting (32 on the CAPQL hopper tasks); mechanism probes use 3 seeds at 60k steps. Seeds are shared across conditions, so all statistics are paired. Our buffer implementation is verified as a drop-in replacement: with all interventions disabled it reproduces CAPQL's internal buffer exactly, in storage order and sampling behavior. To check that the budget is not itself driving the verdicts, we re-ran the two environments whose baselines are
flat at 150k for 400k steps: both then learn ($+34.9$ [$+19.9,+59.0$] and $+5.6$ [$+3.4,+9.7$] EUM) and relabeling remains harmless on them ($d = -0.06$, $+0.04$), so the classification is a property of those tasks rather than of the horizon.

\subsection{Statistical methodology}
The primary metric is final EUM; HV is secondary (as $\mathbf{r}_{\text{ref}}$ dependence introduces a free parameter), and an HV result is only reported when its sign is stable across a $2 \times 2$ grid of reference points and normalization schemes. Effect sizes are Cohen's $d$ with 95\% BCa bootstrap confidence intervals (10K resamples, paired by seed), corrected by Holm-Bonferroni \cite{holm1979simple} within each family of simultaneous tests (one family per intervention across the nine environments). Claims of no effect require equivalence under TOST with a standardized margin $\delta_d = 0.5$, and a claim of harm additionally requires that the baseline itself learns on that task (final versus early EUM, CI excluding zero).

Alongside EUM and HV, which are scalarized and so cannot on their own distinguish a collapsed policy from a diverse one achieving similar value, we report the coverage statistics $C$ and APM (Section \ref{subsec:metrics}), computed from each run's saved final Pareto front on the same preference grid $W$ used for EUM.

\section{Results}
\label{sec:results}

We test the three claims of Section~\ref{sec:pcc} in order. We first verify that the relabeling operator itself is well posed (Section~\ref{subsec:clip}), then establish that relabeling degrades learning and that its incidence is algorithm- and environment-dependent, rule out the obvious confounds, identify the harm as preference coverage collapse, and finally show that mixed relabeling repairs it at a single fixed coefficient.

\subsection{The Standard Operator Silently Ignores Cost Objectives}
\label{subsec:clip}

The apparent effect of relabeling depends first on how the achieved outcome is normalized into a preference. The standard operator normalizes $\text{clip}(\text{achieved}, 0, \infty)$, so an objective that is negative by convention contributes zero: on the four suite tasks carrying a control or energy cost, 86--100\% of relabels degenerate to a corner or uniform vector, confounding the measured harm with the agent being trained to ignore an objective. A sign-robust operator (\texttt{her\_scaled}, per-objective min-max normalization) removes the artifact where the clip binds. The effect on \texttt{ant-2obj} goes from $d = -1.41$ to $+0.04$. This leaves the harm unchanged where it does not, as on \texttt{hopper-2obj} (Appendix B of the supplementary material). All results below use the sign-robust operator, so any remaining effect reflects the relabeling principle rather than its implementation.

\subsection{Well-Posed Relabeling Degrades Learning}
\label{subsec:main-result}

\begin{figure}[t]
\centering
\includegraphics[width=\columnwidth]{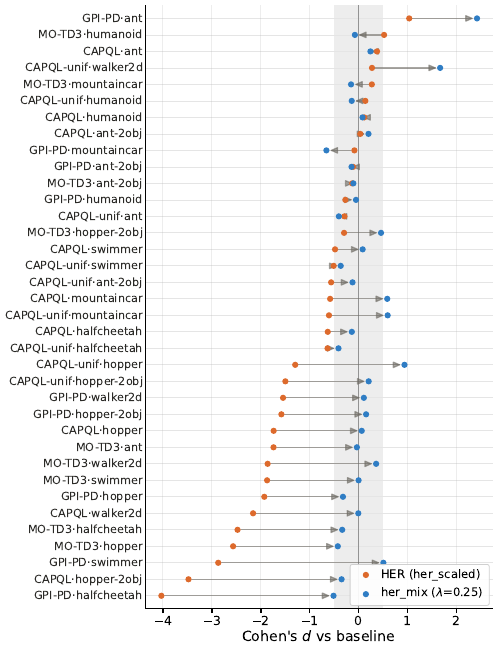}
\caption{Effect of preference relabeling (orange) and of mixed relabeling at $\lambda = 0.25$
(blue) versus baseline, Cohen's $d$ on final EUM, across all 36 algorithm-environment settings,
sorted by the relabeling effect. The grey band marks $|d| < 0.5$; arrows connect the two conditions
per setting.}
\label{fig:validate}
\end{figure}

Even with a well-defined operator, relabeling harms performance in over half of the settings where it has any effect. Across 36 settings, it harms 19 ($d \leq -0.5$ with CI excluding zero), leaves 16 statistically indistinguishable from baseline, and improves exactly one,  with effects ranging from $d = -4.03$ (GPI-PD/halfcheetah) and $+1.03$ (GPI-PD/ant); Figure~\ref{fig:validate} (orange) reports every setting.

Harm has a two-level structure. A core is algorithm-independent: the early-terminating bipeds:
three-objective \texttt{hopper} degrades under all four algorithms ($d$ from $-1.30$ to $-2.56$),
\texttt{hopper-2obj} and \texttt{walker2d} under three. Outside that core, incidence tracks the algorithm: GPI-PD is severely harmed on halfcheetah ($-4.03$) where CAPQL is near-null, MO-TD3 is harmed on ant where GPI-PD improves, and humanoid is unaffected throughout. Whether relabeling hurts is a joint property of the environment's failure structure and of how the algorithm consumes its
conditioning preference. The improvements are not weak-baseline artifacts: every verdict passes the learning gate, and on GPI-PD ant relabeling lifts final EUM from $108$ to $124$.

\subsection{The Harm is intrinsic to Relabeling }

Three families of alternative explanation fail to account for the effect (statistics in Appendices C and D of the supplementary material). Prioritized sampling is not responsible: PER alone is indistinguishable from baseline, and the PER $\times$ HER interaction is identifiable on only two of eight environments, pointing in opposite directions. Buffer structure is not either: re-tested at 16 seeds (256 runs, MDE $|d| \approx 1.02$), no storage, partitioning, eviction, or capacity variant harms performance. Nor is it about which transitions get relabeled: swimmer and halfcheetah never terminate early yet are among the most harmed ($d = -2.87$, $-4.03$), and de-noising the target with return-to-go recovers only 7\% of the loss.

\subsection{The Harm Is Preference Coverage Collapse}
\label{subsec:mechanism}



\begin{figure}[t]
\centering
\includegraphics[width=\columnwidth]{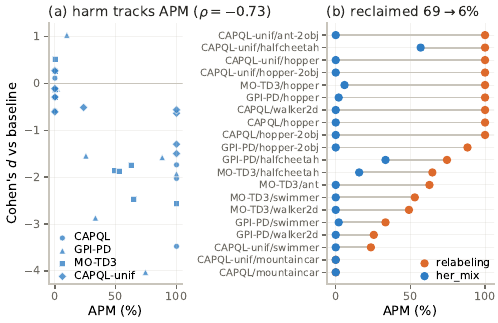}
\caption{\textbf{Coverage collapse is general, and mixing reclaims it.} \textbf{(a)} harm versus
abandoned preference mass across all 32 settings with usable fronts; marker shape denotes the
algorithm. \textbf{(b)} on the harmed settings, \texttt{her\_mix} at $\lambda = 0.25$ returns the
abandoned mass from a mean of $69\%$ to $6\%$.}
\label{fig:pcc-general}
\end{figure}

Measured with the statistics of Section~\ref{subsec:metrics}, coverage collapses under relabeling. Effective coverage $C$ (Equation~\ref{eq:coverage-number}) falls on 22 of 32 settings with usable fronts (mean $2.6\!\rightarrow\!1.9$ effective trade-offs), and on the harmed settings relabeling abandons a mean APM of $69\%$ of the preference simplex. The harm tracks APM (Figure~\ref{fig:pcc-general}a; Spearman $\rho=-0.73$, $p<10^{-5}$) far more tightly than $C$ ($+0.40$):
APM flags the most-harmed CAPQL/hopper-2obj setting (APM $=100\%$), which $C$ misses since that front keeps its diversity while losing value everywhere, and it reads near-zero on the one setting where relabeling helps (APM $=10\%$ on GPI-PD/ant), where collapsing onto a good corner is benign specialization rather than failure. Mixed relabeling reclaims the coverage: on the harmed settings it cuts APM from $69\%$ to $6\%$ (Figure~\ref{fig:pcc-general}b) and restores $87\%$ of the lost $C$, matching its recovery of EUM (Section~\ref{subsec:fix}).

Decomposing performance by preferences shows why. Recomputing per-preference utility from saved Pareto fronts, three of four decisive harmed settings preserve utility near the direction the agent actually achieved and degrade steeply elsewhere.; relabeling concentrated training on the corner reached, abandoning the rest of the simplex. This is predominantly a uniform downward sift rather than seed-level collapse (which dominated on only four of eight harmed settings).

\begin{figure}[t]
\centering
\includegraphics[width=\columnwidth]{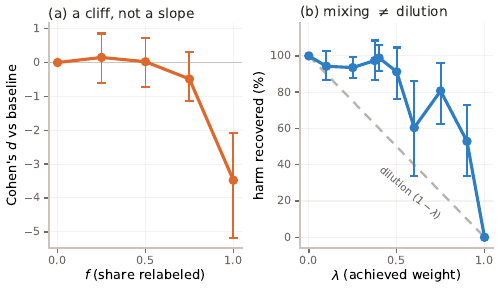}
\caption{\textbf{The failure requires near-total substitution.} Both panels are CAPQL $\times$
hopper-2obj. \textbf{(a)} relabeling each transition with probability $f$: performance holds until
$f \rightarrow 1$. \textbf{(b)} varying the mixing weight $\lambda$: recovery stays near-complete
well past the $(1-\lambda)$ dilution line. $f$ and $\lambda$ are different interventions. Bars are
95\% bootstrap CIs.}
\label{fig:substitution}
\end{figure}

The dose-response is a cliff, not a slope (Figure~\ref{fig:substitution}a). Relabeling transitions independently with probability $f$ on hopper-2obj leaves performance intact through $f = 0.5$ ($d = +0.15$ at $f{=}0.25$, $+0.02$ at $f{=}0.5$), degrades at $f = 0.75$ ($-0.49$), collapsing at $f = 1$ ($-3.48$). A minority of surviving collected preferences suffices to hold the critic's coverage together; the failure mode requires near-total substitution.

Noisy value targets are not the mechanism; instrumenting both critic families, the relabeled TD error is smaller than the collected-conditioning TD error 
in 34 of 36 settings, and its ratio correlates positively with harm ($\rho = +0.36$, CI $[+0.01, +0.68]$), the opposite of what a noisy-target account predicts; our prespecified criterion for that account fails on all 7 environments where algorithms diverge. The harmed critics are the ones that fit the achieved direction best. This rules out repairs that filter high-error relabels, and motivates repairing the conditioning distribution instead.
 
\subsection{Mixed Relabeling Restores Coverage}
\label{subsec:fix}

We evaluate \texttt{her\_mix} (Equation~\ref{eq:hermix}, $\lambda = 0.25$ fixed in Section~\ref{subsec:hermix}) alongside a TD-gated control the coverage account predicts should underperform, on settings spanning all three regimes against criteria fixed in advance (Table~\ref{tab:fix}). Mixed relabeling recovers 87 to 117\% of the harm on all five harmed settings, preserves and amplifies the single benefit (ant: $d = +1.03$ under full relabeling, $+2.42$ under mixing, $+4.46$ at $\lambda = 0.5$), and leaves the neutral control unchanged ($+0.08$, CI $[-0.51, +0.60]$). The TD gate recovers only 62\% on halfcheetah and swimmer and fails its criterion on both, despite accepting 75 to 88\% of relabels, as predicted.

Mixing is not merely attenuated relabeling. Under pure attenuation, recovery would scale as $(1 - \lambda)$ (Figure~\ref{fig:substitution}b); measured recovery is instead flat at 94 to 103\% for all $\lambda \leq 0.5$ on hopper-2obj and collapses only as $\lambda \rightarrow 1$, mirroring the dose-response cliff. On ant, mixing outperforms both of its endpoints, which no interpolation of their outcomes can produce; the blend reaches trade-offs neither pure strategy trains. The recovery is also stable in time, exceeding 86\% over the second half of training on every harmed setting.

\begin{table}[htpb!]
\centering
\small
\setlength{\tabcolsep}{4pt}
\caption{Mixed relabeling on the seven pre-registered settings, spanning the three regimes. $d$ is Cohen's $d$ on final EUM against baseline; recovery is $(\text{mixed}-\text{relabel})/(\text{baseline}-\text{relabel})$ with a paired bootstrap CI, so $100\%$ is a full return to baseline and $0\%$ is no better than relabeling. All arms use the same fixed $\lambda = 0.25$.}
\label{tab:fix}
\resizebox{\columnwidth}{!}{%
\begin{tabular}{llrrr}
\toprule
\textbf{Setting} & \textbf{Regime} & \textbf{$d$ relabel} & \textbf{$d$ mixed} & \textbf{Recovery} \\
\midrule
CAPQL hopper-2obj & harmed & $-3.48$ & $-0.35$ & $94\%\ [87, 101]$ \\
CAPQL hopper & harmed & $-1.74$ & $+0.06$ & $103\%\ [83, 130]$ \\
GPI-PD hopper & harmed & $-1.93$ & $-0.32$ & $95\%\ [85, 109]$ \\
GPI-PD halfcheetah & harmed & $-4.03$ & $-0.51$ & $87\%\ [73, 103]$ \\
GPI-PD swimmer & harmed & $-2.87$ & $+0.51$ & $117\%\ [100, 137]$ \\
GPI-PD ant & helped & $+1.03$ & $+2.42$ & benefit kept \\
CAPQL humanoid & neutral & $+0.12$ & $+0.08$ & unchanged \\
\bottomrule
\end{tabular}%
}
\end{table}

\subsection{Generalisation across the suite}
\label{subsec:validation}

Finally, we deploy the same untuned $\lambda = 0.25$ on every remaining setting of the study (465 additional runs). Figure~\ref{fig:validate} shows the outcome on all 36 settings: 32 satisfy their criterion. Harm is recovered on 16 of 19 harmed settings, across all four algorithms; the three exceptions are settings where relabeling's own effect is small ($|d| \approx 0.6$), so the recovery ratio is noisy even though mixing's absolute effect already lies within the equivalence band. The helped setting keeps its benefit, and 15 of 16 neutral settings are unaffected, with a single regression on GPI-PD mountaincar ($d = -0.66$, CI $[-1.07, -0.36]$). A single fixed coefficient thus converts preference relabeling from an intervention that can cost four standard deviations into one that is neutral to beneficial across two critic families and nine environments.

\section{Discussion, Limitations, and Future Work}
\label{sec:discussion}

Our results reframe HER-style preference relabeling from a free sample-efficiency tool into an
intervention whose safety depends on an under-examined precondition: that an achieved trade-off is a
valid substitute for a requested one. That precondition frequently fails, and where it does the
substitution is not merely noisy but actively harmful. Any relabeling scheme that substitutes an
\emph{achieved} outcome for a \emph{requested} one should therefore be treated as suspect by
default, not assumed safe by analogy to HER's single-objective success.

The failure is repairable, and diagnosing it takes more than coverage. \texttt{her\_mix} removes the
harm without per-environment tuning, and on the one setting where relabeling helps it exceeds both
the pure-relabel and the pure-baseline condition: the gain is access to trade-offs neither endpoint
reaches, not attenuation. Yet effective coverage $C$ correlates only weakly with harm
($\rho = +0.40$) where APM tracks it tightly ($\rho = -0.73$), and APM correctly reads near-zero on
that same helped setting, where collapsing onto a corner is legitimate specialization. Measuring how
much of the preference space an agent covers is not the same as measuring how well it performs
across that space, and a diagnostic built only on the first can be misled in either direction.

Three limitations bound these claims. All results are on MO-MuJoCo with two or three objectives and four algorithm/sampler combinations over two critic backbones, so we do not claim generality across off-policy MORL: discrete-action benchmarks, more than three objectives, and multi-policy or model-based algorithms remain untested. \texttt{her\_mix} is a strong default rather than a guarantee: untuned at $\lambda = 0.25$ it recovers or preserves 32 of 36 verdicts, with one regression (GPI-PD/\texttt{mountaincar}) and partial recovery on the largest original harm (GPI-PD/\texttt{halfcheetah}), both consistent with an account predicting that mixing helps wherever
collapse drives the harm, not that one coefficient closes every gap. And our sign-robust operator resolves the specific clip failure we identified without ruling out other ways an operator could misrepresent an achieved trade-off.

Each points to a next step: establishing whether the collapse mechanism is general to preference-conditioned critics or specific to the single-policy off-policy setting studied here; an adaptive $\lambda$, annealed with training progress or driven by APM as an online proxy, which is already computable from checkpointed fronts; and a lightweight precondition check (flagging majority-negative objectives, say) that catches this failure automatically rather than by dedicated analysis.

\clearpage

\bibliography{aaai2027}

\appendix

\section{Algorithm Details}
\label{app:algorithms}

All four algorithms condition the critic on the preference vector $\mathbf{w}$ and differ in the critic update rule and in how training preferences are drawn. All run with library-default hyperparameters; the replay buffer is the only component we vary.
We also release the full implementation (the swappable buffer, both algorithm adapters, the run manifests, and the
analysis scripts that regenerate every number and figure reported here).

\textbf{CAPQL} \cite{lu2023multi} builds on Soft Actor-Critic \cite{haarnoja2018soft}, conditioning
both actor and critic on $\mathbf{w}$: the actor outputs $\pi(a \mid s, \mathbf{w})$ and the critic
estimates $Q(s, a, \mathbf{w})$. Because linear scalarization concentrates optimal policies at
vertices of the achievable value-function set, CAPQL augments the reward with an entropy term, which
makes that set strictly convex and the preference-to-value mapping continuous. Training preferences
are drawn from an angle cone around the diagonal.

\textbf{GPI-PD} \cite{alegre2023sample} builds on TD3 \cite{fujimoto2018addressing}, likewise
conditioning the critic on $\mathbf{w}$. Rather than shaping a single conditioned critic's output,
it acts by evaluating a support set $\mathcal{M}$ of previously learned weight vectors under the
query preference and taking the best resulting action,
\[
\pi^{\text{GPI}}(s; \mathbf{w}) \in \arg\max_a \max_{\mathbf{w}' \in \mathcal{M}}
Q(s, a, \mathbf{w}') \cdot \mathbf{w},
\]
which is guaranteed to perform at least as well as any single policy in the set. The same support
set supplies the conditioning preferences its critic is trained on.

\textbf{MO-TD3} and \textbf{CAPQL-uniform} are diagnostic ablations that each swap one component.
MO-TD3 pairs the TD3 backbone with CAPQL's cone sampler; CAPQL-uniform pairs the SAC backbone with
uniform sampling over the simplex. Together with CAPQL and GPI-PD they form a $2 \times 2$ grid over
$\{\text{update rule}\} \times \{\text{sampler}\}$, which is what lets us attribute an effect to the
backbone, to the sampler, or to their combination.

\section{The Clip Artifact on Cost Objectives}
\label{app:clip}

Four MO-MuJoCo tasks carry an objective that is a pure control or energy cost and is therefore
negative by construction. The standard \texttt{her\_achieved} operator normalizes
$\text{clip}(\hat{\mathbf{o}}, 0, \infty)$, so on those tasks the clip zeroes that entry on
effectively every transition: the relabeled preference can never point toward the cost objective and degenerates to a corner or uniform vector.

Table~\ref{tab:clip} reports the fraction of genuine (non-degenerate) relabels alongside the effect
of relabeling under both operators. Replacing the clip with per-objective min-max normalization
(\texttt{her\_scaled}) restores informative relabels: the effect on \texttt{ant-2obj} vanishes
entirely and becomes statistically equivalent to baseline, and it shrinks by more than half on
swimmer and halfcheetah. Walker2d retains substantial harm, and \texttt{hopper-2obj}, where the clip never binds, is unchanged. The artifact therefore explains the cost-objective tasks but not the phenomenon itself, which is why all results in the main text use the sign-robust operator.

\begin{table}[htpb!]
\centering
\small
\setlength{\tabcolsep}{5pt}
\caption{Effect of relabeling (Cohen's $d$ on final EUM, CAPQL) under the standard clip-based
operator and the sign-robust operator. Genuine \% is the fraction of relabels that are not corner or
uniform degenerate under the clip.}
\label{tab:clip}
\resizebox{\columnwidth}{!}{%
\begin{tabular}{lccc}
\toprule
\textbf{Environment} & \textbf{Genuine \%} & \textbf{$d$ (clip)} & \textbf{$d$ (sign-robust)} \\
\midrule
ant-2obj & 14\% & $-1.41$ & $+0.04$ \\
swimmer & 0\% & $-1.50$ & $-0.48$ \\
halfcheetah & 0\% & $-1.80$ & $-0.63$ \\
walker2d & 5\% & $-12.0^\dagger$ & $-2.16$ \\
\midrule
hopper-2obj (control) & 92\%+ & $-2.01$ & $-3.48$ \\
\bottomrule
\end{tabular}%
}
\vspace{2pt}
\\ \footnotesize{$^\dagger$Inflated by low pooled variance.}
\end{table}

\section{Alternative Explanations for the Harm}
\label{app:confounds}

The main text states three verdicts on alternative explanations for the harm; this appendix gives
the underlying statistics. Unless noted, the environment is \texttt{hopper-2obj} with termination
on, 16 seeds, and effect sizes are Cohen's $d$ on final EUM with paired BCa intervals.

\subsection{Prioritized sampling}

PER on its own is close to inert, ranging from $d = +0.05$ to $-0.43$ across the suite. For the
interaction we report the difference-in-differences contrast on raw EUM,
$I = (\text{PER}{+}\text{HER}) - \text{PER} - \text{HER} + \text{baseline}$, so that no verdict
rests on a ratio of two noisy effect sizes; $I < 0$ means PER amplifies the harm.
Table~\ref{tab:per-her} gives the per-environment result.

\begin{table}[htpb!]
\centering
\small
\setlength{\tabcolsep}{4pt}
\caption{PER $\times$ HER interaction, 16 seeds per cell, on the \texttt{her\_achieved} cells (the
only ones with a complete $2\times2$). The interaction is identifiable on 2 of 8 environments, and
those two point in opposite directions.}
\label{tab:per-her}
\resizebox{\columnwidth}{!}{%
\begin{tabular}{lrl}
\toprule
\textbf{Environment} & \textbf{$I$ (raw EUM) [95\% CI]} & \textbf{Reading} \\
\midrule
swimmer      & $-1.6$ [$-3.3$, $-0.0$]    & PER amplifies \\
walker2d     & $+38.6$ [$+14.2$, $+64.7$] & PER blunts \\
\midrule
hopper-2obj  & $-5.7$ [$-38.4$, $+26.1$]  & not identifiable \\
hopper       & $+13.2$ [$-7.3$, $+33.6$]  & not identifiable \\
halfcheetah  & $-15.9$ [$-55.6$, $+23.5$] & not identifiable \\
humanoid     & $-0.9$ [$-67.0$, $+66.1$]  & not identifiable \\
ant-2obj     & $-10.4$ [$-72.0$, $+53.1$] & not identifiable \\
ant          & $-1.4$ [$-16.3$, $+12.3$]  & not identifiable \\
\bottomrule
\end{tabular}%
}
\end{table}

On six of eight environments the interval crosses zero, so the interaction is not identifiable at 16
seeds and no directional claim should be read from the point estimate. Where it is identifiable the
two environments disagree in sign. PER is therefore neither an independent source of harm nor a consistent amplifier of it.

\subsection{Early termination}

Whether hopper's early termination amplifies the harm is a $\text{termination} \times \text{HER}$
interaction, which we report as the difference-in-differences $J$ on raw EUM under the primary
sign-robust operator. With termination on, relabeling costs $d = -2.86$ [$-5.46$, $-1.12$]; with it
disabled, $d = -0.51$ [$-0.78$, $+0.11$], giving $J = -105.3$ EUM [$-139.9$, $-66.0$], $p < 0.001$.
Termination therefore amplifies the harm substantially, but does not create it: swimmer and
halfcheetah never terminate early and are among the most harmed settings under the TD3-family
algorithms ($d = -2.87$ and $-4.03$).

The same interaction is not identifiable under the \texttt{her\_achieved} clip
($J = -14.4$ [$-59.1$, $+34.9$], $p = 0.54$), because the clip behaves differently again when
termination is disabled. This is a further illustration of why the operator has to be fixed before
any mechanism question can be asked.

\subsection{Relabel-direction noise and episode length}

Two interventions test whether the harm is an artefact of \emph{which} transitions are relabeled
rather than of relabeling itself. Replacing the one-step achieved direction with the far less noisy
return-to-go direction (\texttt{her\_future}) recovers 7\% of the loss. Suppressing relabels from
episodes shorter than $L_{\min}$ steps recovers a third at $L_{\min} = 50$, rising to 61\% at
$L_{\min} = 200$ while relabeling is still active on 76\% of transitions
(Table~\ref{tab:targeted}).

\begin{table}[htpb!]
\centering
\small
\setlength{\tabcolsep}{5pt}
\caption{Targeted fixes on \texttt{hopper-2obj} (\texttt{her\_achieved}, 16 seeds). Recovery is
$(\text{arm} - \text{HER}) / (\text{baseline} - \text{HER})$.}
\label{tab:targeted}
\begin{tabular}{lrrr}
\toprule
\textbf{Buffer configuration} & \textbf{EUM} & \textbf{$d$} & \textbf{Recovery} \\
\midrule
baseline                          & 178.3 & ---     & --- \\
PER, no HER                       & 165.8 & $-0.43$ & --- \\
HER                               & 81.7  & $-2.97$ & --- \\
PER $+$ HER                       & 76.1  & $-3.55$ & --- \\
\midrule
HER $+$ return-to-go              & 88.4  & $-2.46$ & 7\% \\
HER $+$ length filter ($L{=}25$)  & 80.5  & ---     & $-1$\% \\
HER $+$ length filter ($L{=}50$)  & 113.2 & $-2.58$ & 33\% \\
HER $+$ length filter ($L{=}100$) & 114.1 & ---     & 34\% \\
HER $+$ length filter ($L{=}200$) & 140.3 & ---     & 61\% \\
\bottomrule
\end{tabular}
\end{table}

Short failed episodes are thus disproportionately responsible, which is consistent with the coverage
account: they are exactly the transitions whose achieved direction is degenerate. But no filter
closes the gap, and de-noising the target barely moves it, so neither the noise of the relabel
target nor the population of episodes it is drawn from is the mechanism.

\section{Structural Buffer Design at Conclusion-Grade Power}
\label{app:structural}

The structural knobs were originally screened at 4 seeds, where the minimum detectable effect is
$|d| \approx 2.38$ and no equivalence verdict is possible. We re-ran them at 16 seeds (256 runs),
which brings the minimum detectable effect to $|d| \approx 1.02$, and tested equivalence with TOST
at the preregistered margin $\delta_d = 0.5$. Two environments were used: \texttt{hopper-2obj},
which carries the harm, and \texttt{swimmer}, where the 4-seed screen produced the noisiest
estimates; Table~\ref{tab:structural} reports every cell. Capacity at $1\times$ and $4\times$ the
training budget is omitted: the buffer never
fills at 150k steps, so those settings are mechanically identical to baseline rather than
empirically equivalent to it.

\begin{table}[htpb!]
\centering
\small
\setlength{\tabcolsep}{4pt}
\caption{Structural knobs versus baseline, 16 seeds per cell, CAPQL. $p$ is Holm-corrected across
the 16 tests. No cell shows harm; the single Holm-significant cell is a benefit that does not
replicate across environments.}
\label{tab:structural}
\resizebox{\columnwidth}{!}{%
\begin{tabular}{llrrl}
\toprule
\textbf{Knob} & \textbf{Env} & \textbf{$d$ [95\% CI]} & \textbf{$p$} & \textbf{Verdict} \\
\midrule
storage scalarized      & hopper-2obj & $+0.13$ [$-0.20,+0.57$] & 1.000 & equivalent \\
storage scalarized      & swimmer     & $-0.23$ [$-0.78,+0.35$] & 1.000 & inconclusive \\
partition per-objective & hopper-2obj & $+0.67$ [$+0.22,+1.17$] & 0.022 & \textbf{benefit} \\
partition per-objective & swimmer     & $-0.34$ [$-0.91,+0.11$] & 1.000 & inconclusive \\
partition per-pref.     & hopper-2obj & $-0.09$ [$-0.53,+0.38$] & 1.000 & equivalent \\
partition per-pref.     & swimmer     & $+0.31$ [$-0.21,+0.97$] & 1.000 & inconclusive \\
eviction FIFO           & hopper-2obj & $+0.21$ [$-0.17,+0.56$] & 1.000 & equivalent \\
eviction FIFO           & swimmer     & $-0.22$ [$-0.85,+0.27$] & 1.000 & inconclusive \\
eviction reservoir      & hopper-2obj & $-0.10$ [$-0.50,+0.24$] & 1.000 & equivalent \\
eviction reservoir      & swimmer     & $-0.06$ [$-0.58,+0.31$] & 1.000 & equivalent \\
eviction stratified     & hopper-2obj & $+0.24$ [$-0.01,+0.62$] & 0.450 & inconclusive \\
eviction stratified     & swimmer     & $-0.04$ [$-0.61,+0.60$] & 1.000 & inconclusive \\
capacity $0.25\times$   & hopper-2obj & $+0.21$ [$-0.17,+0.56$] & 1.000 & equivalent \\
capacity $0.25\times$   & swimmer     & $-0.22$ [$-0.85,+0.27$] & 1.000 & inconclusive \\
uniform relabeling      & hopper-2obj & $-0.15$ [$-0.75,+0.25$] & 1.000 & inconclusive \\
uniform relabeling      & swimmer     & $+0.12$ [$-0.49,+0.80$] & 1.000 & inconclusive \\
\bottomrule
\end{tabular}%
}
\end{table}

Six of sixteen cells are equivalent to baseline at $\delta_d = 0.5$ and nine are inconclusive at
that margin, which is a statement about the margin rather than about the effects: every inconclusive
interval is well inside $\pm 1$, an order of magnitude short of relabeling's $d = -4.03$. The single Holm-significant cell is per-objective partitioning helping on \texttt{hopper-2obj}
($d = +0.67$, $p = 0.022$), which reverses sign on \texttt{swimmer} ($-0.34$) and so does not
support a general recommendation. We report it rather than omit it, but it does not bear on the
paper's claim: no structural choice we tested harms performance, and none approaches the magnitude
of the relabeling effect.

\section{Per-Setting Validation of the Fixed Coefficient}
\label{app:validate}

The main paper reports that one untuned $\lambda = 0.25$ satisfies 32 of the 36
algorithm-environment settings, and shows the outcome as a forest plot. Table~\ref{tab:validate}
gives the underlying per-setting numbers. Settings are grouped by the role HER's own effect assigns
them --- harmed, helped, or neutral --- and sorted within each group by that effect. Recovery is
$(\text{mixed} - \text{relabel}) / (\text{baseline} - \text{relabel})$, so $100\%$ is a full return
to baseline; a checkmark marks the pre-registered criterion being met (harmed: recovery $\geq 70\%$
or TOST-equivalent to baseline; helped: $d \geq +0.5$ retained; neutral: not significantly harmed).

\begin{table}[htpb!]
\centering
\small
\setlength{\tabcolsep}{3.5pt}
\caption{Per-setting validation of \texttt{her\_mix} at a single fixed $\lambda = 0.25$. Blocks are
harmed, helped, then neutral settings.}
\label{tab:validate}
\resizebox{\columnwidth}{!}{%
\begin{tabular}{llrrlrc}
\toprule
\textbf{Algorithm} & \textbf{Environment} & \textbf{$n$} & \textbf{$d$ HER} & \textbf{$d$ \texttt{her\_mix}} & \textbf{Recovery} & \\
\midrule
GPI-PD & halfcheetah & 16 & $-4.03$ & $-0.51$ [-1.08,+0.30] & +87\% & $\checkmark$ \\
CAPQL & hopper-2obj & 32 & $-3.48$ & $-0.35$ [-0.79,+0.06] & +94\% & $\checkmark$ \\
GPI-PD & swimmer & 16 & $-2.87$ & $+0.51$ [-0.06,+1.12] & +117\% & $\checkmark$ \\
MO-TD3 & hopper & 16 & $-2.56$ & $-0.42$ [-1.19,+0.41] & +84\% & $\checkmark$ \\
MO-TD3 & halfcheetah & 16 & $-2.47$ & $-0.33$ [-0.97,+0.33] & +81\% & $\checkmark$ \\
CAPQL & walker2d & 16 & $-2.16$ & $-0.01$ [-0.72,+0.74] & +100\% & $\checkmark$ \\
GPI-PD & hopper & 16 & $-1.93$ & $-0.32$ [-1.21,+0.68] & +95\% & $\checkmark$ \\
MO-TD3 & swimmer & 16 & $-1.87$ & $+0.00$ [-0.47,+0.60] & +100\% & $\checkmark$ \\
MO-TD3 & walker2d & 16 & $-1.86$ & $+0.36$ [-0.26,+1.04] & +120\% & $\checkmark$ \\
MO-TD3 & ant & 16 & $-1.74$ & $-0.03$ [-0.65,+0.70] & +98\% & $\checkmark$ \\
CAPQL & hopper & 32 & $-1.74$ & $+0.06$ [-0.43,+0.58] & +103\% & $\checkmark$ \\
GPI-PD & hopper-2obj & 16 & $-1.58$ & $+0.15$ [-0.61,+0.84] & +105\% & $\checkmark$ \\
GPI-PD & walker2d & 16 & $-1.54$ & $+0.11$ [-0.79,+0.94] & +108\% & $\checkmark$ \\
CAPQL-unif & hopper-2obj & 16 & $-1.50$ & $+0.21$ [-0.33,+0.67] & +107\% & $\checkmark$ \\
CAPQL-unif & hopper & 16 & $-1.30$ & $+0.94$ [+0.33,+1.59] & +153\% & $\checkmark$ \\
CAPQL & halfcheetah & 16 & $-0.63$ & $-0.14$ [-0.70,+0.19] & +79\% &  \\
CAPQL-unif & mountaincarcontinuous-v0 & 16 & $-0.60$ & $+0.59$ [-0.01,+1.06] & +115\% & $\checkmark$ \\
CAPQL & mountaincarcontinuous-v0 & 16 & $-0.58$ & $+0.59$ [-0.07,+1.14] & +111\% &  \\
CAPQL-unif & ant-2obj & 16 & $-0.56$ & $-0.12$ [-0.67,+0.34] & +73\% &  \\
\midrule
GPI-PD & ant & 16 & $+1.03$ & $+2.42$ [+1.80,+3.17] & --- & $\checkmark$ \\
\midrule
CAPQL-unif & halfcheetah & 16 & $-0.63$ & $-0.41$ [-0.96,+0.03] & --- & $\checkmark$ \\
CAPQL-unif & swimmer & 16 & $-0.51$ & $-0.36$ [-0.92,+0.13] & --- & $\checkmark$ \\
CAPQL & swimmer & 16 & $-0.48$ & $+0.08$ [-0.50,+0.66] & --- & $\checkmark$ \\
MO-TD3 & hopper-2obj & 16 & $-0.30$ & $+0.46$ [-0.36,+1.09] & --- & $\checkmark$ \\
CAPQL-unif & ant & 16 & $-0.29$ & $-0.40$ [-1.95,+0.41] & --- & $\checkmark$ \\
GPI-PD & humanoid & 16 & $-0.27$ & $-0.05$ [-0.87,+0.83] & --- & $\checkmark$ \\
MO-TD3 & ant-2obj & 16 & $-0.13$ & $-0.11$ [-0.83,+0.70] & --- & $\checkmark$ \\
GPI-PD & ant-2obj & 16 & $-0.11$ & $-0.14$ [-0.58,+0.49] & --- & $\checkmark$ \\
GPI-PD & mountaincarcontinuous-v0 & 16 & $-0.09$ & $-0.66$ [-1.07,-0.36] & --- &  \\
CAPQL & ant-2obj & 16 & $+0.04$ & $+0.20$ [-0.29,+0.54] & --- & $\checkmark$ \\
CAPQL & humanoid & 31 & $+0.12$ & $+0.08$ [-0.51,+0.60] & --- & $\checkmark$ \\
CAPQL-unif & humanoid & 15 & $+0.14$ & $-0.14$ [-0.89,+0.73] & --- & $\checkmark$ \\
MO-TD3 & mountaincarcontinuous-v0 & 16 & $+0.27$ & $-0.15$ [-0.79,+0.35] & --- & $\checkmark$ \\
CAPQL-unif & walker2d & 16 & $+0.27$ & $+1.67$ [+1.07,+2.51] & --- & $\checkmark$ \\
CAPQL & ant & 16 & $+0.38$ & $+0.24$ [-0.82,+0.91] & --- & $\checkmark$ \\
MO-TD3 & humanoid & 16 & $+0.52$ & $-0.08$ [-0.67,+0.42] & --- & $\checkmark$ \\
\bottomrule
\end{tabular}%
}
\end{table}

Sixteen of the nineteen harmed settings are recovered. The three exceptions are settings where
relabeling's own effect is small ($|d| \approx 0.6$), so the recovery ratio is noisy even though
mixing's absolute effect already lies inside the equivalence band. The single helped setting keeps
its benefit, and fifteen of sixteen neutral settings are unaffected, the exception being a
regression on GPI-PD/\texttt{mountaincar} ($d = -0.66$, CI $[-1.07, -0.36]$).


\end{document}